\documentclass[11pt,a4paper]{article}
\usepackage[dvipsnames]{xcolor}
\usepackage{tikz}
\usepackage[hyperref]{acl2021}
\usepackage{multirow}
\usepackage{times}
\usepackage{latexsym}

\usepackage{graphicx}
\usepackage{amsmath}
\usepackage{amssymb}
\usepackage{caption}
\usepackage{subcaption}
\usepackage{boldline}
\usepackage[normalem]{ulem}
\usepackage{microtype}

\aclfinalcopy % Uncomment this line for the final submission
\def\aclpaperid{2751} %  Enter the acl Paper ID here

\title{Representing Syntax and Composition with Geometric Transformations}

\author{Lorenzo Bertolini~~~Julie Weeds~~~David Weir~~~Qiwei Peng\\
  University of Sussex \\
  Brighton, UK \\
  \texttt{\{l.bertolini, j.e.weeds, d.j.weir, qiwei.peng\}@sussex.ac.uk} \\}

\date{}

\begin{document}
\maketitle
\begin{abstract}
The exploitation of syntactic graphs (SyGs) as a word's context has been shown to be beneficial for distributional semantic models (DSMs), both at the level of individual word representations and in deriving phrasal representations via composition. However, notwithstanding the potential performance benefit, the syntactically-aware DSMs proposed to date have huge numbers of parameters (compared to conventional DSMs) and suffer from data sparsity. Furthermore, the encoding of the SyG links (i.e., the syntactic relations) has been largely limited to linear maps. The knowledge graphs' literature, on the other hand, has proposed light-weight models employing different geometric transformations (GTs) to encode edges in a knowledge graph (KG). Our work explores the possibility of adopting this family of models to encode SyGs. Furthermore, we investigate which GT better encodes syntactic relations, so that these representations can be used to enhance phrase-level composition via syntactic contextualisation.
\end{abstract}

\section{Introduction}\label{sec:intro}
Representing words in terms of their syntactic co-occurrences has been long proposed, both for count-based~\citep{Pado_2007,Weir_2016}, and neural~\citep{hermann-blunsom-2013-role,Levy_2014,komninos-manandhar-2016-dependency, czarnowska-etal-2019-words, vashishth-etal-2019-incorporating} models of word meaning.  Tested on benchmark word similarity tasks, such models often perform favourably to models based on proximal co-occurrence, particularly when the similarity or substitutability of two words is considered rather than their relatedness~\cite{Levy_2014}.  However, the real promise of distributional models based on syntactic rather than proximal co-occurrence, is the potential for carrying out syntax-sensitive composition.  For example, in the Anchored Packed Tree (APT) model~\citep{Weir_2016} lexemes, phrases, and sentences are represented as collections of typed occurrences, and composition is carried out by contextualising each element in its syntactic role.  This leads to syntax-sensitive representations for phrases.  For example, \emph{glass window} and \emph{window glass} have different representations due to the different syntactic roles played by each constituent.

Alongside count-based models, a variety of neural ones have been proposed to encode syntactic structure, focusing on different depths of the graph~\citep{Levy_2014,komninos-manandhar-2016-dependency, marcheggiani-titov-2017-encoding, vashishth-etal-2019-incorporating,emerson-2020-autoencoding}).  Of particular note here, ~\citet{Levy_2014}~and~\citet{komninos-manandhar-2016-dependency}~each proposed models (DEP and EXT, respectively) which learn from local dependency relations, by extending the Skip-Gram with Negative sampling (SGNS) architecture from \texttt{word2vec}~\cite{Mikolov_2013}. Given a tuple of \textit{(target, context)} words, e.g. \textit{(rain,like)}, a standard SGNS model can be trained to encode the probability of it being a true or a randomly sampled tuple. DEP and EXT, on the other hand, make use of both standard and syntactically contextualised tuples e.g., ~(\textit{rain\_$\overline{dobj}$}, $like$)\footnote{$\overline{dobj}$ indicating the inverse of the $dobj$ relation}. Whilst DEP was tested solely on word similarity tasks,~\citet{komninos-manandhar-2016-dependency} applied large neural architectures to sentence level tasks and were thus able to demonstrate a positive impact of applying an additive composition strategy to syntax-aware representations. 

There is of course an explosion in the number of parameters to be learnt in both DEP and EXT due to the many possible word-relation combinations which form the target vocabulary for these models (see Table \ref{tab:mdl_prm}). A possible solution, proposed by~\citet{czarnowska-etal-2019-words}, is the Dependency Matrix (DM) model which uses linear maps in the form of square matrices to encode relations. Here, the training objective is changed from predicting \textit{(target, context)} pairs to \textit{(target, relation, context)} triples, e.g., \textit{(rain,$\overline{dobj}$,like)}. This model produced comparable results with DEP and EXT at the word level.  Furthermore, compositional experiments on short phrases, specifically relative clauses, produced encouraging results when using the learned transformations. Yet, despite considerably reducing the number of parameters, this model still makes use of large word spaces and the square linear map is still costly to train.

\begin{table}[htb!]
\centering
% \small
{%
\begin{tabular}{l|c|}
\multicolumn{1}{c|}{Model} & Learnable Parameters \\ \hline
DEP                         & 223M               \\
DM                         & 51.6M                \\
MuRE                       & 21.5M                \\
RotE                       & 21.5M                \\
RefE                       & 21.5M                \\
AttE                       & 21.6M               
\end{tabular}%
}
\caption{\label{tab:mdl_prm}Learnable parameters for each model, given the same word (72k) and relation (88) vocabularies from the ~\texttt{text8} (parsed) corpus, and vector size of $n$=300.}
\end{table}

The reformulation of the SGNS objective introduced by DM (i.e., moving from \textit{(target,context)} tuples to \textit{(target,relation,context)} triples) closely resembles a common practice in the knowledge graphs (KGs) literature (e.g. \cite{Trouillion_complex,mure_murp,chami-etal-2020-rote}). Here, large, mainly factual, graphs are fed to neural models in the form of \textit{(head,relation,tail)}. Compared to the syntactically-aware DSMs discussed above, many of the models proposed to encode KGs make use of a substantially lower number of parameters to encode both word and relations, as shown in Table \ref{tab:mdl_prm}.  Furthermore, in order to represent the heterogeneous types of relations in KGs, researchers have experimented with models based on different types of geometric transformations (GTs). These include, but are not limited to, stretch \cite{mure_murp}, rotation \cite{sun2018rotate, chami-etal-2020-rote}, reflection \cite{chami-etal-2020-rote} and attention \cite{chami-etal-2020-rote}. However, in the KG literature, limited attention has been paid to the compositional nature of phrases. Single-token oriented vocabularies (where \textit{New York} is represented by \textit{New\_York}), used in most KGs, work well for real-world entities, such as people or cities, but are problematic when considering compositional phrases such as $\textit{small cake}$. As discussed by \citet{toutanova-etal-2015-representing}, treating these phrases in the same way forces the vocabulary to grow immensely, and prevents the model from reasoning over new phrases in a compositional fashion. Hence, developing successful composition strategies is of interest to the KG community as well as more widely in Natural Language Inference (NLI).

Given the success that DM and other models have obtained in modelling syntax and syntactically driven composition, we propose to overcome the parameter and word-relation vocabulary problems by using GT models to encode syntactic graphs. We focus our investigation on four state of the art models from the knowledge-graphs literature, namely MuRE~\cite{mure_murp}, and the three GTs-based models proposed by~\citet{chami-etal-2020-rote}: RotE, RefE and AttE. Despite the simplicity, MuRE has obtained competitive results, when compared to more complex models~\cite{chami-etal-2020-rote}). Rotation has been used to model composition of relation representations~\cite{sun2018rotate}. Attention has been frequently proposed as a plausible mechanism for composition (e.g.~\citet{arad2018compositional,NIPS2019_8845,yin-etal-2020-sentibert,russin-etal-2020-compositional}), whilst reflection is relatively under-studied~\cite{chami-etal-2020-rote}. Furthermore, as discussed in Section~\ref{sec:methods}, these models allow for an interesting comparison, as they can be grouped into three categories: tail modifiers (DM), head modifiers (RotE, RefE, AttE), and full modifiers (MuRE).  Hence, we explore some of the transformational properties required to enable the successful encoding of syntactic relations, where success is defined in terms of their potential to support phrasal composition.  

Our contributions are as follows. First, we show how lighter-weight models based on GTs can be used to encode both word and syntactic relations, frequently outperforming DM both in word similarity and compositional benchmarks. Second, for each model, we propose a tailored composition strategy, based on syntactic contextualisation of one (or more) of the phrase constituents. We hence show how to exploit the learned syntactic representations for composition, by comparing syntax-driven strategies for composition with simple addition.  Third, we provide an analysis of which type of GTs better encode relations for syntactic contextualisation and enhanced composition.

\section{Related Work}\label{sec:rel_wrk}
\label{sec:realted_work}
Knowledge graphs are complex data structures where nodes are concepts or entities (usually content words like \emph{dog} or \emph{Campari}) and edges are relations (e.g. \texttt{is\_a}, \texttt{produced\_in}) connecting entities to one another (e.g. \emph{dog} \texttt{is\_a } \emph{mammal}, \emph{Campari} \texttt{produced\_in} \emph{Italy}). Table~\ref{tab:ds_stats} reports the number of distinct entities, relations and triples for three of the most investigated KGs, namely, FB15k-237~\cite{toutanova-chen-2015-observed} YAGO3-10~\cite{Mahdisoltani2015YAGO3AK}, and WN18RR~\citep{wn188rr_1_dettmers2018conve}, as well as a syntactic graph (SyG) constructed from the parsed corpus \texttt{text8}. The way these graphs are structured can vary significantly.~\citet{chami-etal-2020-rote} showed how, among the presented KGs, only WN18RR has a significantly hierarchical structure.

\begin{table}[htb!]
\centering
% \small
\resizebox{\columnwidth}{!}
{%
\begin{tabular}{l|cccc|}
Dataset    & entities & relations & triples   & graph type \\ \hline
WNRR18                          & 31k      & 11        & 87k       & KG         \\
FB15k-237                       & 15k      & 237       & 272k      & KG         \\
YAGO3-10                        & 123k     & 33        & 1M        & KG         \\
\texttt{text8} & 72k      & 88        & 12M$^{*}$ & SyG       
\end{tabular}%
}
\caption{\label{tab:ds_stats}Statistics for the training splits of different datasets (* number of unique items, with observed repetitions, items raise to 18M).}
\end{table}

Research on models for representing KGs has mainly focused on the ability to predict new connections between existing nodes. To overcome the problem of testing items that do not occur in the training set, many models have adopted negative sampling (NS) strategies in the training phase.  The vocabulary of KG datasets is also largely single-token oriented. Models able to handle multi-token items have been proposed \citep{toutanova-etal-2015-representing,  toutanova-etal-2016-compositional, sun2018rotate}, but they focus on the composition of relations rather than entities, e.g., how a complex relation such as \texttt{married\_to:son\_of} might be split into multiple constituents and composed. Also relevant,~\citet{toutanova-chen-2015-observed} showed how syntax-augmented triples extracted from documents (e.g. $(Obama,\texttt{nsubj:born\_in:obj}, USA)$) can be beneficial for KGs models, but did not investigate representing syntax or composition via embeddings.

Previous works (e.g.~\citep{marcheggiani-titov-2017-encoding, vashishth-etal-2019-incorporating}) showed how SyGs could be encoded via graph convolutional networks (GCN)~\cite{Kipf2017SemiSupervisedCW}. These large models are able to encode larger graphs (up to the sentence level), via sequences of convolutions along the edges of the graph. Such convolutions are frequently relation-specific and are also encoded via square matrices. 

\section{Theoretical Approach}\label{sec:th_app}
\label{sec:methods}
In both the semantic (KG) and syntactic (SyG) domain, the starting point is typically a dataset $D$ of positive triples $(h,r,t)$, with $h,t$ $\in$ $V=\{1,..,|V|\}$ and $r \in$ $R=\{1,..,|R|\}$, where $V$ and $R$ are the sets of the indexes for the vocabulary of entities / words and relations, respectively. In both domains, the shared goals are: i) map entities $v$ $\in$ $V$ to embeddings $e_{v}$ where $e$ $\in$ $\mathbb{R^{|\text{V}| \times \textit{n}}}$, $n$ being the dimensionality of the vectors; ii) map relations $r$ $\in$ $R$ in one – or more – space $\mathbb{R^{|\text{R}| \times *}}$.  In this work, we focus on constructing a syntactic dataset of positive training triples from a corpus as in~\citet{czarnowska-etal-2019-words}. All of the models we investigate rely on a negative sampling mechanism that generates a dataset $D^{'}$ of false triples. Each model was presented in its own original work with a tailored way to generate $D^{'}$. Unless otherwise stated, we make use of the original mechanism. 

As already discussed, we are interested in both word level and compositional level evaluation. Testing at the word level, e.g., using word similarity benchmarks, simply requires extraction of the word embeddings. Compositional tests, on the other hand, also require syntactic analysis of the phrase and extraction and application of the relation embeddings. The first step, is to generate a parsed version of the phrase. For example, syntactic analysis of the phrase $pour~tea$ will produce the root-as-head (Rh) $(h,r,t)$ triple $(pour, dobj, tea)$, and the root-as-tail (Rt) $(h,r,t)$ triple $(tea, \overline{dobj}, pour)$.  Such duplicity of representations was handled in DM by obtaining both representations and then summing the cosine similarities obtained when comparing each of the two representations with a given target.  Whilst reasonably effective in the DM evaluation, this does not provide a single phrase-level representation and would become unwieldy for longer phrases and sentences.  ~\citet{Weir_2016} argued in favour of considering the syntactic root as the main element of any multi-token linguistic item. In our example, to compare \emph{pour tea} with \emph{drink water}, this would require us to consider the syntactic root in the context of its dependent i.e., how similar is the verb \emph{pour} when contextualised by the direct object \emph{tea} to the verb \emph{drink} when contextualised by  the \texttt{direct\_object} \emph{water}? In models which modify the head of the triple (e.g., \cite{chami-etal-2020-rote}, this would correspond to using the root-as-tail (Rt) analysis of the phrase. Here, we compare the two strategies empirically.  Further, inspired by the growing success of (very large) bi-directional models such as ELMo~\cite{Peters_2018} and BERT~\cite{Devlin_2018} and also by recent evidence from the neuroscientific literature~\cite{Mollica2020, Fedorenko_2020}, suggesting that sentence processing strongly relies on identifying and composing smaller units of meaning, such as phrases, regardless of order of their constituents, we also propose a third compositional strategy which is bi-directional in nature. Here, the phrase-level representation is the sum of the root-as-head and the root-as-tail representations, making it more agnostic to the direction of the relation as well as the word order.  However, phrases with different structures such as \emph{glass window} and \emph{window glass} will still have different representations due to the different roles played by each word in each relation.

In summary, we propose and investigate three different syntax-aware (\emph{syn}) composition strategies: \emph{syn}-Rh and \emph{syn}-Rt, different solely in where the root is placed in the $(head,relation,tail)$ triple; and \emph{syn}-BiD (for bi-directional), constructed by adding the representations obtained by \emph{syn}-Rh and \emph{syn}-Rt. We now describe in detail the models investigated, together with our tailored \emph{syn} composition strategy for each of them. 

\paragraph{DM} This model is an extension of SGNS, where a linear map, in the form of a \textit{n$\times$n} matrix, projects a word from the context space $(e')$ into the target space $(e)$, as in Equation \ref{dm_u}:
\begin{equation}
    u = e_h^T \cdot (W_r e^{'}_{t})
\label{dm_u}
\end{equation}
where $e,e' \in \mathbb{R^{|\text{V}| \times \textit{n}}}$, and $W$ $\in$ $\mathbb{R^{|\text{R}| \times \textit{n} \times \textit{n}}}$.  Since the tail word is projected into the space occupied by the head word, we refer to this model as a tail-modifier.  ~\textit{u} is then used to compute standard SGNS loss (Equation \ref{dm_loss}):

\begin{equation}
% \small
\sum_{(h,r,t)~\in~D}\log\sigma(u)~+~\sum_{(h,r,t)~\in~D^{'}}\log\sigma(-u)
\label{dm_loss}
\end{equation}
Phrase representations will be constructed following our three syntactic composition strategies. As a baseline, common to all models, we use addition (\emph{add}) of the queried head and tail entities embeddings, as in Equation~\ref{add_cmp}~\footnote{This corresponds to simple-sum composition in the original work by ~\citet{czarnowska-etal-2019-words}.}:
\begin{equation}
    e_{add} = e_h + e_{t}
\label{add_cmp}
\end{equation}
We propose \emph{syn} composition for the DM model to be obtained via $u$ (Equation~\ref{dm_u}), as in Equation~\ref{dm_ph}:
\begin{equation}
    e_{syn} = e_h + (W_r e^{'}_{t})
\label{dm_ph}
\end{equation}

\paragraph{MuRE} This architecture falls into the family of translation models~\cite{chami-etal-2020-rote}. Here,  both the entities go through a transformation and so we refer to this model as a full-modifier. The tail entity is shifted with a translation (i.e. offset), and a stretch, in the form of a~\textit{n$\times$n} diagonal matrix, is applied to the head entity. Embeddings are then fed to a distance function $d(x,y)=\left\|x-y\right\|$ and the model minimises the Bernoulli negative log-likelihood loss, using Equation~\ref{mure_d}, to estimate the probability of the triple being from $D$:
\begin{equation}
% \small
p(h,r,l) = \sigma(-d(W_{r}e_{h},e_{t}+w_{r})^{2}+b_{h}+b_{t})
\label{mure_d}
\end{equation}
Here, $W$ $\in$ $\mathbb{R^{|\text{R}| \times \textit{n} \times \textit{n}}}$ contains $|R|$ diagonal matrices (each corresponding to a relation-specific stretch),  $w$ $\in$ $\mathbb{R^{|\text{R}| \times \textit{n}}}$ hosts $|R|$ translation vectors, and $b$ $\in$ $\mathbb{R^{|\text{V}| \times \textit{n}}}$ the entity biases. Again, additive composition is carried out by adding the queried embedding for the phrase's constituents. Syntactic composition is implemented by adapting the model's score function (Equation~\ref{mure_ph}):
\begin{equation}
    e_{syn} = W_{r}e_{h}+(e_{t}+w_{r})
\label{mure_ph}
\end{equation}
\paragraph{RotE, RefE} These models optimise a full cross-entropy loss. Like MuRE, square distance between two vectors is used as a score function. Unlike the previous model, they apply a Givens rotation (Rot) or reflection (Ref), as defined in~\citet{chami-etal-2020-rote}, and a translation to the head entity.  Thus, we refer to these models as head-modifiers.  Syntactic composition is defined via the score functions in Equations \ref{rote_ph} and \ref{refe_ph}:
\begin{equation}
% \small
    e_{syn} = (\text{Rot}(T_{r})e_{h}+t_{r}) + e_{t}
\label{rote_ph}
\end{equation}
\begin{equation}
% \small
    e_{syn} = (\text{Ref}(F_{r})e_{h}+f_{r}) + e_{t}
\label{refe_ph}
\end{equation}
where $T,F$ $\in$ $\mathbb{R}^{|\text{R}| \times \frac{n}{2}}$ each contain $|R|$ diagonal matrices (each corresponding to a relation-specific Givens rotation or reflection), and $t,f$ $\in$ $\mathbb{R^{|\text{R}| \times \textit{n}}}$ are relation-specific translations.

\paragraph{AttE} Intuitively, AttE is designed to model the contribution of different GTs (in this case just rotation and reflection). This is achieved via a self-attention mechanism. Given two embeddings $x$, $y$, and an attention vector $a$, attention scores are computed via Equation~\ref{atte_aph}:
\begin{equation}
% \small
    (\alpha_x, \alpha_y) = \text{Softmax}(a^{T}x,a^{T}y)
\label{atte_aph}
\end{equation}
These scores are then averaged (Equation~\ref{atte_slfatt}):
\begin{equation}
% \small
    \text{Att}(x, y; a) = (\alpha_{x}x+\alpha_{y}y)
\label{atte_slfatt}
\end{equation}
To actively select the most suitable transformation for a given triple, rotation and reflection are applied to the head-entity embedding (Equation~\ref{qrt_qrf}):
\begin{equation}
% \small
   \text{q}_{\text{Rot}} = \text{Rot}(T_{r})e_{h},~\text{q}_{\text{Ref}} = \text{Ref}(F_{r})e_{h}
\label{qrt_qrf}
\end{equation}
The two representations are than combined using a self attention mechanism (Equation~\ref{atte_Q}):
\begin{equation}
% \small
    Q(h,r) = \text{Att}(\text{q}_{\text{Rot}},\text{q}_{\text{Ref}};a_{r})+p_{r}
\label{atte_Q}
\end{equation}
with $p$ $\in$ $\mathbb{R^{|\text{R}| \times \textit{n}}}$ as the relation-specific translation.
$Q$ and the $e_{t}$ are then used as arguments for $d$ as in Equation~\ref{mure_d}. Syntactically contextualised composition (\emph{syn}) for AttE is implemented via Equation~\ref{atte_ph}:
\begin{equation}
% \small
    e_{syn} = Q(h,r) + e_{t}
\label{atte_ph}
\end{equation}

\section{Experiments}\label{sec:exp}
Our main aim is to investigate the potential of models in terms of constructing high quality word representations and their support for composition. To this end, experiments were carried out with a set of models trained on KGs, and a second set of models trained on SyGs. This allows us to investigate the value of encoding distributional information from SyGs or whether KGs alone might be a sufficient source of data to obtain competitive results. We hypothesise that when using KGs alone: i) word similarity tasks might yield high results; ii) compositional evaluation will yield poor results. As for models trained on SyG, we expect to see: i) a generally improved performance on most tasks, when compared to models trained on KGs; ii) larger models to be penalised across benchmarks and for syntactically-contextualised (\emph{syn}) composition.

\subsection{Experimental setup}
\paragraph{Benchmarks} We divide our quantitative experiments between word similarity and composition tasks. For the word similarity tasks, we focus on SimLex~\citep{Hill_2015}, MEN~\citep{Bruni_2014}, and both similarity (WS\_s) and relatedness (WS\_r) split of the WordSim353~\citep{Finkelstein_2001} datasets. For every word pair, we produce a model's prediction using cosine similarity (CS). We compare model predictions and human judgements using Spearman's $\rho$.

For the compositional investigation, we focus on the~\citet{Mitchell_2010} (ML10) dataset. Items in this benchmark consist of pairs of two-token phrases (e.g. \textit{(pour tea–drink water)}) paired with human judgements on their similarity. Phrases are composed using the four different presented strategies and the obtained representations are compared via CS. Again, CS and human ratings are compared via $\rho$. We selected this benchmark for two main reasons: i) the models' structures lend themselves straightforwardly to syntactically contextualised (\emph{syn}) composition strategies for a two-token item\footnote{\citet{czarnowska-etal-2019-words} proposed a more complex composition strategy, specifically for relative clause phrases which we do not consider here.}; ii) the dataset is pre-split into three syntactic-relation classes (i.e. adjective-nouns (AN), verb-objects (VO) and noun-nouns (NN)) and this division offers an opportunity for a more in-depth investigation on how different models and operations manage to embed different syntactic relations. 

We trained each set of models with three random initialisation, and report the mean and standard error (SE) of the obtained  $\rho$s.

\paragraph{Implementation} For MurE, RotE, RefE and AttE we adapt the original PyTorch code. Since an official release of the DM is not available, we implemented a PyTorch version of the model\footnote{\url{https://github.com/lorenzoscottb/findings_ACL2021}}. 

We trained the first set of GT models on the WN18RR dataset, tuning negative sampling rate (NS), optimiser and learning rate using mean reciprocal rank (MRR) on the development set\footnote{using the dataset's original splits.}. Epochs were kept stable at 50 and $n$ to 300. We focused on WN18RR as YAGO3-10 shares a minimal vocabulary with the selected word-similarity and compositional benchmarks. FB15k-237, on the other hand, has all the entities encrypted. The models obtained from this training set were then evaluated on both word-similarity and compositional tasks (see Table~\ref{tab:wn18_tsk}) to provide a baseline for the SyG models.

\begin{table*}[htb!]
\centering
\resizebox{.9\textwidth}{!}
{%
\begin{tabular}{l|cccc|ccc|}
\textbf{} & Simlex  & MEN     & WS\_s   & WS\_r   & Adjective Nouns & Verb Objects & Noun-Noun    \\ \hline
MuRE      & .38±.01 & .45±.00 & .42±.01 & .21±.03 & .19±.03 & .31±.00 & .13±.01 \\
RotE      & .35±.01 & .54±.00 & .59±.00 & .30±.02 & .18±.03 & .33±.00 & .20±.02 \\
RefE      & .36±.01 & .54±.00 & .57±.00 & .30±.01 & .16±.04 & .37±.01 & .14±.02 \\
AttE      & .36±.01 & .54±.00 & .58±.01 & .29±.00 & .20±.00 & .32±.00 & .18±.00
\end{tabular}%
}
\caption{\label{tab:wn18_tsk}Spearman $\rho$s' (mean ± SE) obtained on all selected benchmarks, for knowledge-graph models trained on WN18RR dataset.}
\end{table*}

A second set of models was trained on the \texttt{text8}\footnote{\url{http://mattmahoney.net/dc/textdata}} corpus, parsed with spaCy~\cite{honnibal-johnson-2015-improved}. Following \citet{czarnowska-etal-2019-words}, minimum item count, epochs, NS, optimiser and learning rate were fine-tuned on SimLex. Hyperparameters are selected from the union of the ones proposed in \cite{mure_murp, czarnowska-etal-2019-words, chami-etal-2020-rote}. All the models share the same number of dimensions, i.e., $n=300$. For a fair comparison, all experiments for this set have been conducted on the vocabulary shared across the models. Final coverage  and best hyperparamenters are reported in Appendix~\ref{subsec:voc_cov} and~\ref{subsec:hyp}. All models were trained using NVIDIA Titan V GPUs.

\subsection{Results}
\paragraph{WN18RR trained models} We begin our quantitative investigation evaluating models from the knowledge graph literature, trained on WN18RR, on all benchmarks. Looking at Table~\ref{tab:wn18_tsk}, we note that these models, compared to models trained on \texttt{text8} or similar distributional models trained on much larger corpora, achieve competitive results on the word similarity benchmarks, especially in the historically challenging SimLex dataset, despite the small vocabulary and training samples.

A possible explanation for these results lies in how entities co-occur in the training data. First of all, WN18RR has a limited vocabulary (see Table~\ref{tab:ds_stats}), and is poorly populated by adjectives. Furthermore, noun and verbs, two part of speech (POS) that frequently co-occur between each other in natural language, here mainly occur within each other (i.e. verb with verb, noun with noun). In few cases, especially for verbs, the co-occurrences are not only limited to the same POS, but interest the very same word. All models perform much worse on the relatedness split of WS-353 than the similarity split. This might be expected, for models trained on WordNet data. As predicted, the performance is generally poor for composition benchmarks. An exception seems to be the VO subset, where models achieve results that, as will be presented shortly, are competitive also for \texttt{text8}-trained models.

\paragraph{Word similarity}
Our motivation for experiments with models trained on \texttt{text8} is to understand whether models previously proposed for representing KGs are competitive with distributional models such as DM in their ability to embed word and syntactic relations. Results for word-similarity are presented in Table~\ref{tab:ww_tsk}.

\begin{table}[htb!]
\centering
\small
\resizebox{\columnwidth}{!}
{%
\begin{tabular}{l|cccc|}
     & Simlex & MEN & WS\_s & WS\_r \\ \hline
DM   & .12±.01  & .60±.01 & .59±.02 & .51±.03 \\ 
MuRE & .17±.01  & .64±.00 & .69±.01 & .58±.00 \\  
RotE & .17±.00  & .64±.00 & .70±.00 & .58±.01 \\ 
RefE & .18±.01  & .63±.01 & .70±.01 & .56±.00 \\ 
AttE & .16±.00  & .61±.00 & .69±.01  & .57±.01
\end{tabular}%
}
\caption{\label{tab:ww_tsk}Spearman $\rho$s' (mean ± SE) obtained on word-word similarity benchmarks, with models trained on \texttt{text8} corpus.}
\end{table}

\begin{table*}[htb!]
\centering
\tiny
\resizebox{.8\textwidth}{!}{%
\begin{tabular}{cl|ccc||c}
\multicolumn{1}{l}{\textbf{}} & \textbf{}    & Adjective Nouns  & Verb Objects     & Noun-Nouns       & Average          \\ \hline
\multirow{4}{*}{DM}           & \textit{add} & .39±.02          & .31±.03          & .43±.03          & .37±.02          \\
                              & \textit{syn}-Rh       & .26±.02          & .18±.02          & .25±.03          & .23±.02          \\
                              & \textit{syn}-Rt       & .32±.03          & .14±.02          & .20±.02          & .22±.02          \\
                              & \textit{syn}-BiD      & .33±.02          & .14±.02          & .34±.03          & .27±.02          \\ \hline
\multirow{4}{*}{MuRE}         & \textit{add} & .47±.01          & .35±.01          & .40±.00          & .41±.00          \\
                              & \textit{syn}-Rh       & \textbf{.51±.01} & .34±.01          & .44±.01          & .43±.00          \\
                              & \textit{syn}-Rt       & .49±.01          & .36±.01          & .43±.01          & .43±.01          \\
                              & \textit{syn}-BiD      & .49±.01          & .36±.01          & \textbf{.46±.01} & .44±.00          \\ \hline
\multirow{4}{*}{RotE}         & \textit{add} & .49±.00          & .37±.00          & .43±.00          & .43±.00          \\
                              & \textit{syn}-Rh       & .48±.01          & .36±.01          & .41±.01          & .42±.01          \\
                              & \textit{syn}-Rt       & .47±.02          & .35±.01          & .41±.01          & .41±.00          \\
                              & \textit{syn}-BiD      & .49±.00          & .38±.00          & .45±.01          & .44±.00          \\ \hline
\multirow{4}{*}{RefE}         & \textit{add} & .48±.01          & .36±.00          & .43±.01          & .42±.00          \\
                              & \textit{syn}-Rh       & .49±.01          & .36±.01          & .43±.01          & .43±.01          \\
                              & \textit{syn}-Rt       & .48±.01          & .34±.02          & .43±.01          & .42±.01          \\
                              & \textit{syn}-BiD      & .48±.00          & \textbf{.38±.01} & \textbf{.46±.01} & \textbf{.44±.01} \\ \hline
\multirow{4}{*}{AttE}         & \textit{add} & .46±.01          & .35±.01          & .41±.01          & .41±.00          \\
                              & \textit{syn}-Rh       & .47±.01          & .35±.00          & .43±.01          & .41±.01          \\
                              & \textit{syn}-Rt       & .45±.01          & .29±.01          & .43±.01          & .39±.00          \\
                              & \textit{syn}-BiD      & .48±.02          & .36±.00          & .46±.00          & .43±.00         
\end{tabular}%
}
\caption{\label{tab:cmp_tsk_fll}Spearman $\rho$s' (mean ± SE) obtained on~\citet{Mitchell_2010} benchmark, with models trained on \texttt{text8} corpus. Phrasal composition is carried out by element-wise addition (\textit{add}), and the three proposed syntax (\textit{syn}) aware strategies: root as head (\textit{syn}-Rh), root as tail (\textit{syn}-Rt) and bidirectional (\textit{syn}-BiD).~\textbf{Best results for each Phrase Type}.}
\end{table*}
First, scores on SimLex are much lower than: i) those achieved by the KG-trained models; ii) those presented elsewhere for DM in the literature \cite{czarnowska-etal-2019-words}. We note that the corpus we used to train the models is significantly smaller than the one used to train DM by the original authors, and we assume that this, combined with the low frequency of SimLex items in our corpus, is the main reason for these differences. Results for DM on the other word similarity benchmarks are much closer to the performance achieved by the original authors and, on these benchmarks, DM clearly outperforms the baseline of models trained on WN18RR. However, most notably, GT models trained on the same data as DM, not only achieve comparable results to DM, but they almost always outperform it, both in similarity-based and relatedness-based benchmarks. Moreover, DM seems to show the highest variation, especially for WN\_s and WN\_r.

\paragraph{Composition} Table~\ref{tab:cmp_tsk_fll} shows  the results for all \texttt{text8}-trained models on the compositional benchmark.  Again, GT models show competitive results, and generally outperform DM, which fails at improving its performance with \emph{syn} composition. This last evidence is reversed in all other models. That is, they all achieve best performance with one of the syntax-aware composition methods. Looking closer, we can see that, in most cases, the best \emph{syn} method is the bi-directional one, with the exceptions of MUuRE, RotE and RefE's AN phrases. Notably, \emph{syn}-BiD is almost never a mere average of the two representations that originated it. In many cases, and especially for AttE, \emph{syn}-BiD representations produce a significantly larger gain in performance, when compared to both \emph{syn}-Rt and \emph{syn}-Rh. From the single model perspective, the best performing one is RefE. Syntax-aware methods based on reflection always outperform the additive baseline, and also obtained the best score in the average sections, via bi-directional composition. Again, DM is the model showing the highest variation in results. This provides further evidences in favour of the lightweight models taken from the KG literature

\subsection{Statistical Analysis}
All correlations were tested for significance, adopting the Holm correction~\cite{Holm1979ASS} to account for the large number of tests, and we observed no \textit{p}~$<$~.05. As the main interest of our work was the compositional investigation (reported in Table~\ref{tab:cmp_tsk_fll}), a global comparison was conducted to test whether observed differences in correlations were also significant. We adopted a paired two-tail bootstrap analysis~\cite{berg-kirkpatrick-etal-2012-empirical, sogaard-etal-2014-whats, dror-etal-2018-hitchhikers}, performed independently between results from the three seeds. Given the large number of comparisons, a Holm correction was adopted within the same Phrase Type. Results (see~\ref{subsec:Significance} for more details) showed that, among all models, the only one that generated a number of  insignificant differences was DM, mainly pertaining to different strategies for composing NN items.

\begin{figure*}[htb!]
    % \centering
    \begin{subfigure}[b]{0.328\textwidth}
        \includegraphics[width=\textwidth]{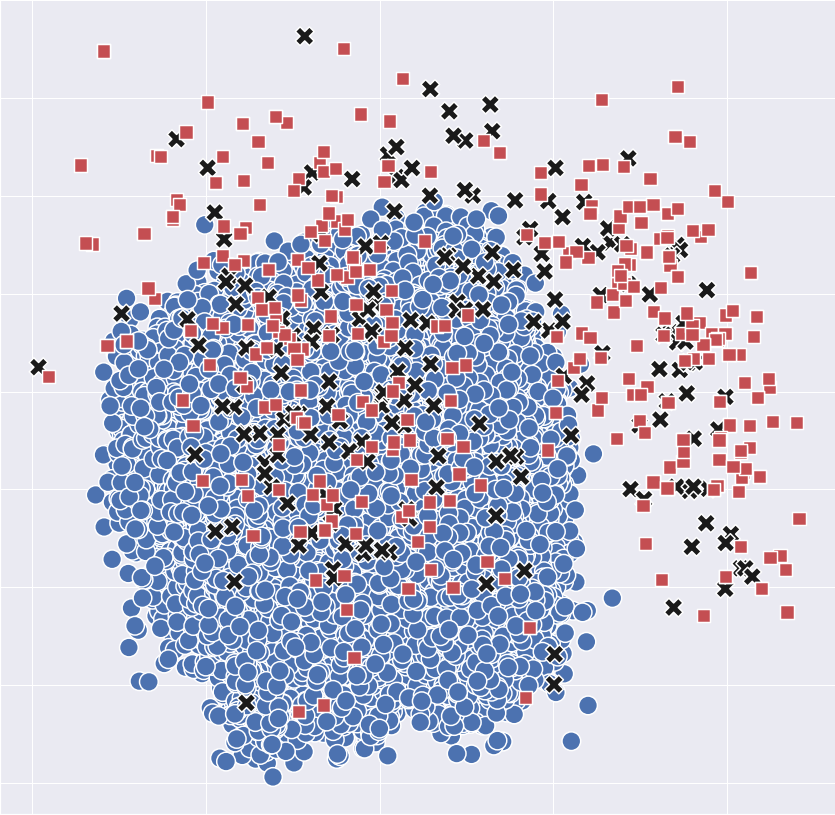}
        \caption{\textit{syn}-Rt}
        \label{fig:refe_rt}
    \end{subfigure}  
    \begin{subfigure}[b]{0.328\textwidth}
        \includegraphics[width=\textwidth]{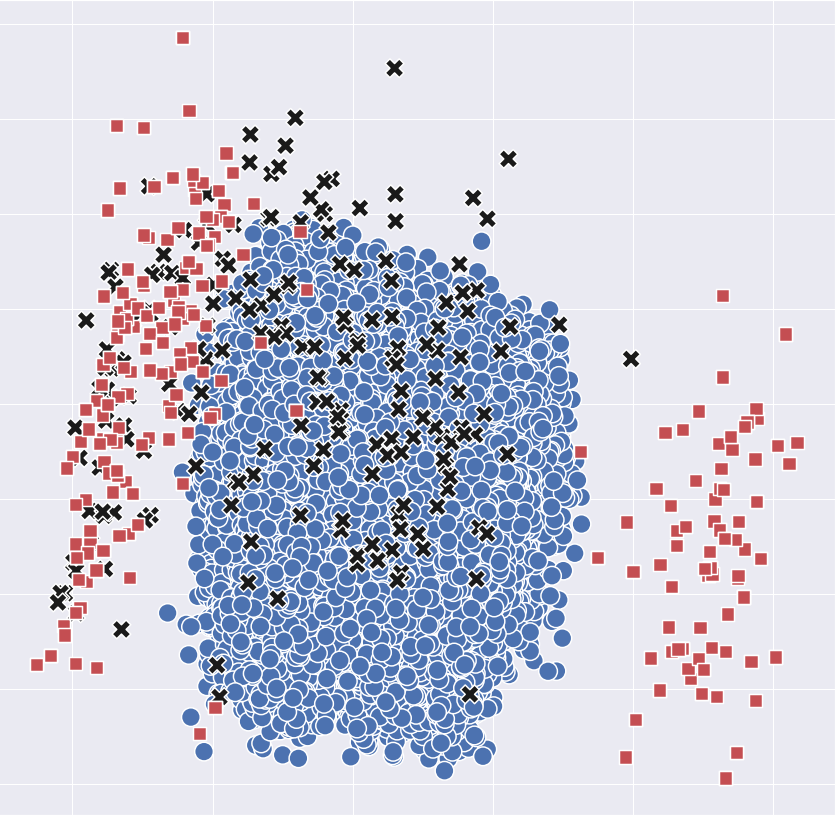}
        \caption{\textit{syn}-Rh}
        \label{fig:refe_rh}
    \end{subfigure}
    \begin{subfigure}[b]{0.328\textwidth}
        \includegraphics[width=\textwidth]{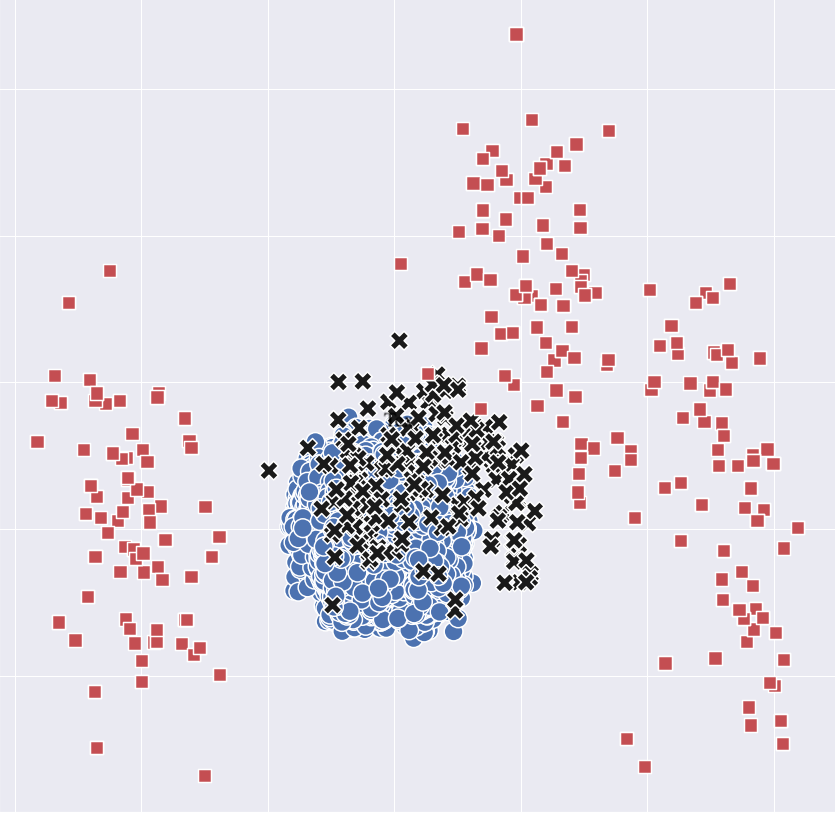}
        \caption{\textit{syn}-BiD}
        \label{fig:refe_bid}
    \end{subfigure}
    \caption{PCA visualisation of RefE vector space. Images show the same word ($\textcolor{NavyBlue}\bullet$) and \textit{add}-composed vectors ($\times$), in the context of representations composed with the four different syntax-aware ($\textcolor{BrickRed}\blacksquare$) composition methods. All composed vectors represent the set of phrases from the~\citet{Mitchell_2010} benchmark.}
    \label{fig:refe_cmp_pca}
\end{figure*}

\subsection{Qualitative Analysis}
\begin{figure}[htb]
    \centering
    \includegraphics[width=.9\columnwidth]{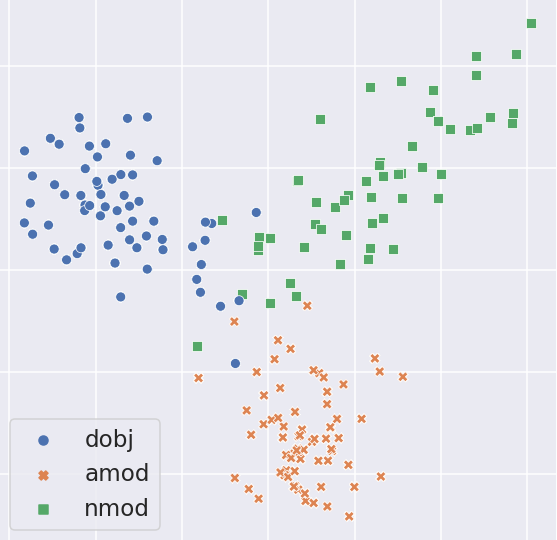}
    \caption{\label{fig:refe_ml10}PCA visualisation of syntactically contextualised root-items from ML10 phrases using RefE and reflection. AN roots are  contextualised using \texttt{amod}($\textcolor{BurntOrange}\times$), VO via \texttt{dobj} ($\textcolor{NavyBlue}\bullet$), NN via \texttt{nmod} ($\textcolor{ForestGreen}\blacksquare$).}
\end{figure}

We now investigate the impact of relation representations on word vectors and composition from a qualitative point of view.  Here, we focus on the model that quantitative tests indicated as the most promising one: RefE. We will start at the word level, looking at syntactically contextualised single words. The interest here, is to see if clear relation-driven clusters can be identified within a reduced space. To do so, we contextualise the set of roots from ML10 (e.g. $amount$ in $vast~amount$), and reduce the dimensions through PCA. Results in Figure~\ref{fig:refe_ml10} suggest that the three syntactic relations adopted for contextualisation (i.e. \texttt{amod}, \texttt{dobj}, \texttt{nmod}) appear to generate as many distinguishable clusters. Despite being limited, these results support evidence for syntactic subspace probed out of mBert~\cite{chi-etal-2020-finding}.

Concluding, we explore how composition strategies behave with respect to the word representations. To do so, we concatenate representations obtained by \textit{add}-composing the set of ML10 items with the full original space, and each syntax-aware strategy separately. The three obtained sets of concatenation (i.e. word–\textit{add}–\textit{syn}-Rt; word–\textit{add}–\textit{syn}-Rh; word–\textit{add}–\textit{syn}-BiD) is then independently reduced to \textit{n}=2 through principal component analysis (PCA). Results are reported in Figure~\ref{fig:refe_cmp_pca}. As it can be observed throughout the three reductions, and mostly in Figure~\ref{fig:refe_bid}, phrase representations obtained via simple addition mainly lie within the perimeter of the word space. A similar pattern is observed in Figure~\ref{fig:refe_rt}, with \emph{syn}-Rt. Phrases composed by using the root as the head of the triple are still fairly close to the word-space perimeter, but tend to abandon its centre. Lastly, Figure~\ref{fig:refe_bid} shows how bi-directional representations lie scattered fairly distant from the word and \emph{add}-composed representations. This last observation is contrary to theories suggesting that representations at every level (word, phrase, sentence, etc..) should lie within the same space (e.g.~\citet{Weir_2016}). However, it may support recent work from neuroscience (e.g. ~\citet{Ding2016CorticalTO}) suggesting that the brain networks processing word, phrases and sentences do not completely overlap.

\section{Discussion}\label{sec:discussion}
Our results strongly suggest that light-weight models presented in the knowledge-graphs literature can be efficiently applied to syntactic-graphs, and be converted to distributional models that are consistently able to make use of the learned word and relation representations to improve semantic phrase-composition. From the model-theoretical point of view, evidence suggests that constraining linear maps with a reflection (together with a non-linear translation) seems to be the most efficient way of encoding syntactic relations. Our quantitative results also contribute to the debates on how sequential language data, or English at least, should be processed and what the role of syntactic information should be. As mentioned in Section~\ref{sec:methods}, the models selected distinguish between being tail (DM), head (RotE, RefE and AttE) and full (MuRE) modifiers. Further, we can change the syntactic focus of any of these models by adopting the \emph{syn}-Rt composition strategy instead of the \emph{syn}-Rh strategy.  However, in our experiments, the head-modifier models (RotE, RefE and AttE) outperformed the tail-modifier and full models (DM and MuRE) and achieved a better results with the \emph{syn}-Rh strategy than the \emph{syn}-Rt strategy, i.e., when the syntactic root of the phrase was taken as the head of the triple rather than as the tail.  In other words, it appears better to contextualise the root and compose with its dependent, which opposes the linguistic arguments put forward by~\citet{Weir_2016}.  However, even more notably, the \emph{syn}-BiD composition strategy, which combines the \emph{syn}-Rh and \emph{syn}-Rt representations, generally gave a further boost to performance. This is further evidence that bi-directional information is more informative than uni-directional information, not just in large neural models such as LSTMs and transformers, and supports recent theory from neuroscience which argues that what is crucial for composition is not the overall structure nor the root, but that we can identify a phrase's constituents and the relation they have \cite{Mollica2020}. Evidence in favour of the fact that composition strongly relies on local dependencies based on syntactic structure was also found by~\citet{saphra-lopez-2020-lstms}. Such work suggests that LSTMs learn to compose following a hierarchical structure, driven by syntax, and that they rely on the learned short sequences to build longer and more reliable ones. Taken altogether, the evidence from different language-related fields is becoming more compelling that syntax and phrase composition should play an important role in the composition of larger units of meaning. 

\section{Conclusions and Further Work}\label{sec:conclusion}
We have shown how GT models previously proposed for encoding KGs can be adapted to encode syntactic information in a distributional model. We have demonstrated the high quality nature of the distributional word representations and the potential for using syntactically-contextualised composition strategies for phrases. In particular, we have demonstrated the competitiveness of lighter-weight GT models when compared to more general models based solely on unconstrained linear maps, such as DM.  Further, our analysis has shown how learned representations for syntactic relations can be efficiently exploited at the word level, transforming a word through part-of-speech related regions of the space, and at the phrase level, generating superior composed representations. Furthermore, we have shown, among the different GTs, reflection seems to be the most promising for encoding syntactic relations. Future work will focus on composition on larger scale, syntactic-relation composition, and whether syntactic and semantic graph can be simultaneously embedded using this framework.

\section*{Acknowledgements}
This research was supported by EPSRC grant no. 2129720: \textit{Composition and Entailment in Distributed Word Representations}. We also thank the anonymous reviewers for their helpful comments, and NVIDIA for the donation of the GPU that supported our work. The first author would also like to thank Gabriele Paveri, for the years of conversations on human language, and constantly doubting the author's ideas.

\bibliographystyle{acl_natbib}
\bibliography{anthology,acl2021}

\clearpage
\appendix

\section{Appendices}
\subsection{Hyperparameters}\label{subsec:hyp}
Table~\ref{tab:tx8_hyp} reports the best obtained hyperparameters for models trained on \texttt{text8} corpus. These are minimum count (MC), negative sample rate (NS), epochs (EP), learning rate (lr), and optimiser (Opt.). For models trained on WN18RR hyperparameter where identical to the ones indicated in the original works, as ide from negative samples (best obtain 10) and epochs, kept at 50, as indicated in the paper. Results Obtained on the WN18RR test split did not significantly differ form the scores reported in the original works. Again, the total set of parameters was obtain by intersecting the ones presented in the models' original papers~\cite{czarnowska-etal-2019-words,mure_murp, chami-etal-2020-rote}.

\begin{table}[htb!]
\centering
\small
\resizebox{\columnwidth}{!}{%
\begin{tabular}{l|llllll}
     & MC & NS & EP & lr & \multicolumn{1}{l|}{Opt.} \\ \hline
DM   & 100       & 20               & 5      & .001      & \multicolumn{1}{l|}{Adam}      \\
MuRE & 0         & 40               & 50     & 50      & \multicolumn{1}{l|}{SGD}      \\
RotE & 0         & 30               & 15     & 50     & \multicolumn{1}{l|}{SGD}      \\
RefE & 0         & 30               & 15     & 50     & \multicolumn{1}{l|}{SGD}      \\
AttE & 0         & 25               & 10     & 50    & \multicolumn{1}{l|}{SGD}                          
\end{tabular}
}
\caption{\label{tab:tx8_hyp}Best hyperparameters for models trained on \texttt{text8} corpus.}
\end{table}

\subsection{Vocabulary Coverage}\label{subsec:voc_cov}
We here present the final coverage for all the benchmarks used for the models trained on the WN18RR (Table~\ref{tab:wn18rr-cov}) and \texttt{text8} (Table~\ref{tab:txt8-cov}) corpora.

\begin{table}[htb!]
\centering
\tiny
\resizebox{\columnwidth}{!}{%
\begin{tabular}{l|l|}
Benchmark            & Coverage  \\ \hline
SimLex               & 726/999   \\
MEN                  & 1544/3000 \\
WS353\_sim           & 152/203   \\
WS353\_rel           & 200/251   \\
ML10 Adjective Nouns & 1836/1944 \\
ML10 Verb Objects    & 1836/1944 \\
ML10 Noun-Nouns      & 1782/1944
\end{tabular}%
}
\caption{\label{tab:txt8-cov}Final coverage of the different datasts' items used for testing models trained on \texttt{text8}.}
\end{table}

\begin{table}[htb!]
\small
\centering
\resizebox{\columnwidth}{!}{%
\begin{tabular}{l|l|}
Benchmark            & Coverage  \\ \hline
SimLex               & 787/999   \\
MEN                  & 1635/3000 \\
WS353\_sim           & 166/203   \\
WS353\_rel           & 200/251   \\
ML10 Adjective Nouns & 648/1944  \\
ML10 Verb Objects    & 1674/1944 \\
ML10 Noun-Nouns      & 1494/1944
\end{tabular}%
}
\caption{\label{tab:wn18rr-cov}Final coverage of the different datasts' items used for testing models trained on WN18RR.}
\end{table}

Note the significantly smaller coverage that models trained on WN18RR show for Adjective Noun phrases on Table~\ref{tab:wn18rr-cov}. Such small coverage is one of the main reason that guided the decision towards not sharing the word vocabulary across models trained on the two different corpora. 

\subsection{Statistical Significance}\label{subsec:Significance}
We here report those Model-Strategy pairs for which the observed differences in the correlation analysis are not statistically significant, according to our bootstrap test.

\begin{table}[htb!]
\centering
\resizebox{\columnwidth}{!}{%
\begin{tabular}{cccc}
\hlineB{3}
\textbf{Phrase Type} & \textbf{Model A} & \textbf{Model B} & \textbf{\textit{p}} \\ \hlineB{3}
NN                   & DM-add           & DM-Rt            & .728      \\
NN                   & DM-Rh            & DM-Rt            & .216      \\ \hline
VO                   & DM-add           & DM-Rh            & .864      \\
NN                   & DM-add           & DM-BiD           & .066      \\ \hline
NN                   & DM-add           & DM-Rh            & .213      \\
NN                   & DM-add           & DM-Rt            & .410      \\
NN                   & DM-Rh            & DM-Rt            & .268      \\
VO                   & DM-add           & DM-Rt            & .147      \\ \hlineB{3}
\end{tabular}%
}
\caption{\label{tab:bootstrap-ml10} Bootstrap analyses results, stratified by different random seeds. \textit{p} values refers to Holm-corrected values.}
\end{table}

\subsection{Single Space DM}\label{subsec:os_md_sec}
\begin{figure*}[htb!]
    \includegraphics[width=\textwidth]{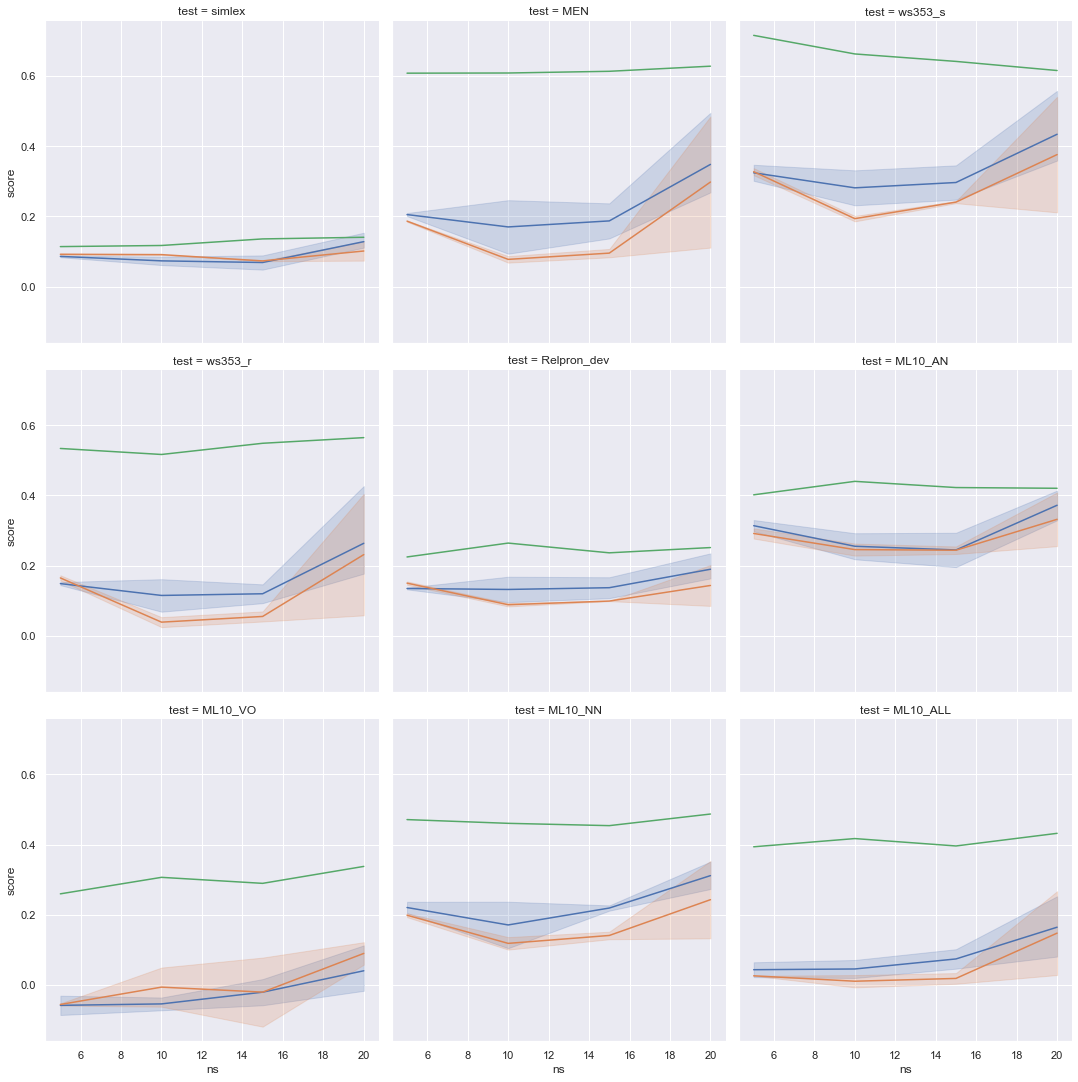}
    \caption{\label{fig:osdm}Comparison of results on all the benchmarks discussed in the paper with a \textcolor{ForestGreen}{DM} model and two single-space version, \textcolor{BurntOrange}{OSDM} and \textcolor{NavyBlue}{FullOSDM}, obtained applying~\citet{zobnin-elistratova-2019-learning} method to the DM. The shaded areas refer to the fact that these models included the extra hyperparameter q.}
\end{figure*}

We are aware that~\citet{zobnin-elistratova-2019-learning} proposed a method to reduce SGNS vector spaces to one, and run a few preliminary experiments adopting this strategy in DM. As presented in Figure~\ref{fig:osdm}, such experiments clearly suggest that DM is superior to the investigated variants.

\end{document}